\documentclass[conference]{IEEEtran}
\IEEEoverridecommandlockouts

\usepackage{cite}
\usepackage{amsmath,amssymb,amsfonts}
\usepackage{hyperref}
\hypersetup{
	colorlinks,
	menucolor=black,
	linkcolor=black,
	citecolor=black,
	urlcolor=blue
}
\usepackage{graphicx}
\usepackage{textcomp}
\usepackage{xcolor}
\def\BibTeX{{\rm B\kern-.05em{\sc i\kern-.025em b}\kern-.08em
    T\kern-.1667em\lower.7ex\hbox{E}\kern-.125emX}}
\begin{document}

\title{Bridging Spiking and Continuous Dynamics in Spiking Neural Networks: A Muscle-Inspired Model \\

}
\author{
\author{\IEEEauthorblockN{Maryam Soltani\IEEEauthorrefmark{1}\IEEEauthorrefmark{2}
		and Philip N. Garner\IEEEauthorrefmark{1}}
	\IEEEauthorblockA{\IEEEauthorrefmark{1}Idiap Research Institute, Martigny, Switzerland}
	\IEEEauthorblockA{\IEEEauthorrefmark{2}\'Ecole Polytechnique F\'ed\'erale de Lausanne (EPFL), Lausanne, Switzerland\\
		Email: maryam.soltani@epfl.ch, phil.garner@idiap.ch}}
}

\maketitle

\begin{abstract}
Transforming the spike trains produced by spiking neural networks (SNNs) into continuous signals through biologically inspired mechanisms remains an open challenge in generative and regression tasks. In biological systems, a similar transformation occurs when muscles integrate neural spikes to produce smooth, coordinated movements. Inspired by this mechanism, we introduce a muscle-response model, a biologically motivated dynamical layer that converts spike trains into continuous representations. Our approach uses second-order dynamics to model the temporal integration of spiking activity. We investigate how the number of spiking neurons and temporal resolution affect the quality of the generated signals in a text-to-speech task, showing that increasing representational capacity has a greater impact than increasing temporal resolution under the evaluated configurations. Furthermore, an analysis of the learned dynamics reveals that, without explicit constraints on the response type, most neurons converge toward non-oscillatory behavior. These results demonstrate the potential of biologically inspired dynamical decoders for bridging spiking neural computation and continuous signal generation.

\end{abstract}


\section{Introduction}
The nervous system communicates almost exclusively through discrete spikes \cite{Gerstner_Kistler_2002}, yet the behavior it produces is continuous. Nowhere is this more evident than in speech. Spike trains from motor neurons drive the muscles of larynx, tongue, and lips \cite{guenther2016speech, bouchard2013speech}, which integrate these discrete events into smooth articulatory movements \cite{heckman2012motorunit} and a rich continuous acoustic signal. The body therefore provides a compelling biological example of a transformation that remains relatively under-explored in artificial spiking systems: reconstructing structured continuous signals from sparse, discrete events through biologically inspired dynamics. Understanding and emulating this spike-to-continuous transformation is a key step if spiking neural networks (SNNs) are to move beyond recognition and toward generation.

SNNs have achieved competitive performance on a growing range of recognition tasks, from image classification \cite{fang2021plif} to speech command recognition \cite{bittar2022,bittar2022surrogate}. In these settings, the network output need only support a discrete decision, and the conversion from spikes to a final answer is comparatively forgiving. Both classification and regression require an interface between spike-based computation and continuous outputs; however, the requirements are more demanding for regression tasks. Unlike classification where the final objective is a discrete decision, regression requires accurate reconstruction of continuous-valued signals while preserving their temporal and structural characteristics. The interface that maps sparse spike trains back to continuous values has received little dedicated attention. This interface, we believe, is not a peripheral detail but a central bottleneck for spike-based generation models.

Beyond their biological appeal, SNNs offer the potential for substantially improved energy efficiency on neuromorphic hardware \cite{neuromorphiccomputing}, making them attractive candidates for future generative models, which are currently computationally expensive. Existing approaches to extracting continuous-valued information from spike trains include neural coding strategies, such as rate and temporal coding \cite{fono2026mathematical}, as well as trainable readouts, including convolutional layers \cite{spikevoice}. While these approaches can effectively recover continuous representations, they do not necessarily incorporate an explicit, biologically inspired dynamical model of how individual spike events are integrated over time.

We instead view spike-to-continuous conversion as a dynamical representation problem: individual spike events must be integrated over time into smooth trajectories without discarding the information carried by their timing. Following the biological principle outlined above, we propose a muscle-response bridge that performs this integration through threshold-less second-order temporal filtering. This formulation is closely related to the Spike Response Model (SRM) \cite{SRM}. In the SRM, the membrane potential of a neuron is written as a weighted superposition of stereotyped kernels, each triggered by an incoming spike. Postsynaptic kernels in this framework typically take a smooth rise-and-decay form, such as a difference of two exponentials. While the SRM describes how incoming spikes shape neuronal membrane potentials, our approach extends the principle of spike-triggered temporal responses to the reconstruction of continuous, high-dimensional outputs from a spiking network.

 Filtering spike-like commands through muscle models also has a precedent in speech prosody. In the generalized command-response (GCR) model, the intonation contour is represented as a superposition of muscle responses to spike commands. Marelli et al.~\cite{marelli2019} embedded trainable second-order filters in a neural network to generate the log-F0 contour, restricting their poles to complex-conjugate pairs and thereby favoring underdamped dynamics. However, their command signals were generated by a conventional recurrent network and encouraged to be sparse through temporal L1 regularization, rather than produced by genuine spiking neurons. Moreover, their model targeted a single low-dimensional prosodic contour. Here, we extend this principle to a genuine spiking network, using trainable second-order dynamics to reconstruct high-dimensional acoustic representations without prescribing a particular damping regime.

To evaluate this bridge, we use speech synthesis as a testbed for continuous reconstruction. Its outputs are high-dimensional, temporally structured, and perceptually sensitive to fine detail. Specifically, we insert a spiking layer followed by the proposed bridge between the decoder and vocoder of FastSpeech2 \cite{fastspeech2}, a widely used text-to-speech model, so that the reconstructed representation can be compared directly against the original non-spiking one. To the best of our knowledge, this is the first work to use a muscle-inspired dynamical layer as the spike-to-continuous interface of an SNN. Our main contributions are:
(1) We propose a biologically inspired spike-to-continuous
interface based on trainable, thresholdless second-order
dynamics that integrates spike trains into
continuous acoustic representations.
(2) We systematically investigate the effects of temporal
resolution and representational capacity on reconstruction
quality, demonstrating that both are important, with
representational capacity having a greater impact under
the evaluated configurations. 
(3) We show that, with sufficient representational capacity, the spiking pathway fully recovers the degradation introduced by the spiking bottleneck, reaching an objective quality comparable to the original non-spiking FastSpeech2 output. We further show that, with increased temporal resolution and capacity, the proposed spiking pathway outperforms a variant using a conventional convolutional readout at the baseline configuration.
(4) We show that, without explicitly constraining the
damping regime, the learned second-order responses
predominantly converge to the overdamped regime,
suggesting that non-oscillatory dynamics are
favored by the speech reconstruction objective.
Together, these findings provide architectural and
dynamical design principles for extending SNNs
toward continuous regression and generative tasks.

\section{Background: Spiking Neuron Model}
\label{sec:background}

SNNs, as opposed to conventional artificial neural networks, use spikes to process information, allowing for energy-efficient computing and the native handling of sequential and temporal data \cite{majumdar2025spiking}. Various models have been proposed to describe spiking neuron dynamics. Among them, the leaky integrate-and-fire (LIF) model originally proposed by Lapicque \cite{lapicque1907recherches} is one of the most widely used neuron models due to its simplicity and computational efficiency. The adaptive leaky integrate-and-fire (AdLIF) model extends the conventional LIF model by incorporating an adaptation mechanism, enabling it to capture biologically relevant dynamics such as spike-frequency adaptation (SFA) \cite{Gerstner_Kistler_2002}. In this study, we employ the AdLIF neuron model, as previous work has demonstrated the benefits of incorporating adaptation compared with conventional LIF neurons, particularly for processing temporally complex signals such as speech \cite{baronig2025}.

\subsection{Leaky Integrate-and-Fire Neuron}
\label{sec:lif}

The LIF dynamics are given by
\begin{equation}
	\tau_m \frac{du}{dt} = -u + R\,I(t),
	\label{eq:lif}
\end{equation}
where $u$ is the membrane potential, $\tau_m = RC$ is the membrane time constant, $R$ is the membrane resistance, and $I(t)$ is the input current. A spike is emitted when the membrane potential reaches a threshold $\vartheta_{\mathrm{th}}$, after which the potential is reset to $u_r$. The spiking train can be expressed in continuous time as $s(t) = \sum_f \delta(t - t^f)$, where $t^f$ denotes the firing times.

\subsection{Adaptive LIF Neuron}
\label{sec:adlif}

To capture SFA, an additional adaptation current is introduced. Using a continuous-time formulation with Dirac impulses avoids explicit reset rules and preserves computational efficiency compared to detailed models such as Hodgkin--Huxley \cite{hodgkin}. After parameter reduction, the model becomes \cite{sparse}:
\begin{align}
	\tau_u \dot{u}(t) &= -u(t) - w(t) + I(t) - \tau_u \sum_f \delta(t - t^f), \label{eq:adlif_u}\\
	\tau_w \dot{w}(t) &= a\,u(t) - w(t) + \tau_w b \sum_f \delta(t - t^f), \label{eq:adlif_w}
\end{align}
where $w(t)$ represents the adaptation current, and $\tau_u$ and $\tau_w$ denote the membrane and adaptation time constants, respectively. This formulation provides a favorable trade-off between computational complexity and biological realism.

The AdLIF neuron is thus characterized by four parameters: the membrane and adaptation time constants $\tau_u$ and $\tau_w$, and the adaptation parameters $a$ and $b$. For numerical implementation, the continuous-time dynamics are discretized using a first-order exponential integration scheme. Defining
\begin{equation}
	\alpha = \exp\!\left(-\frac{\Delta t}{\tau_u}\right), \qquad
	\beta  = \exp\!\left(-\frac{\Delta t}{\tau_w}\right),
	\label{eq:alphabeta}
\end{equation}
the resulting discrete-time dynamics are given by
\begin{align}
	u_t &= \alpha\,(u_{t-1} - s_{t-1}) + (1-\alpha)\,(I_t - w_{t-1}), \label{eq:adlif_disc_u}\\
	w_t &= \beta\,(w_{t-1} + b\,s_{t-1}) + (1-\beta)\,a\,u_{t-1}, \label{eq:adlif_disc_w}\\
	s_t &= \mathbb{I}(u_t \geq 1), \label{eq:adlif_disc_s}
\end{align}
where $I_t$ is the input and in discrete time, the emitted spike is represented by the binary variable \(s_t\in\{0,1\}\). In \eqref{eq:adlif_disc_s}, the threshold is normalized to $\vartheta_{\mathrm{th}} = 1$, and the reset is implemented by subtraction through the term $-s_{t-1}$ in \eqref{eq:adlif_disc_u}, the discrete counterpart of the Dirac term in \eqref{eq:adlif_u}. All four parameters are learned during training and constrained to the biologically plausible ranges reported in \cite{Gerstner_Kistler_2002, sparse, augustin}, namely $\tau_u \in [1, 25]$~ms, $\tau_w \in [30, 450]$~ms, $a \in [-0.5, 5]$, and $b \in [0, 2]$.

\subsection{Training with Surrogate Gradients}
\label{sec:surrogate}

The non-differentiable nature of spike generation in \eqref{eq:adlif_disc_s} makes it impossible to directly apply gradient-based optimization methods. To address this problem, surrogate gradient methods have been widely used \cite{neftci2019surrogate}, in which the derivative of the step function is replaced by a smooth approximation during the backward pass. In this work, we use a boxcar surrogate function \cite{bittar2022surrogate}.

\section{Proposed Muscle-Response Model}
\label{sec:muscle}

Biological motor systems provide a natural example of spike-to-continuous transformation. Discrete neural commands are integrated by the neuromuscular system into smooth, non-oscillatory rise-and-decay responses. Inspired by this principle, we introduce a muscle-response model, a dynamical readout that transforms sparse spike trains into continuous representations through second-order temporal filtering. In the following, we first motivate the second-order formulation and derive the model from the AdLIF neuron (Section~\ref{sec:muscle_formulation}), and then analyze its damping regimes, which we later use to examine whether the learned dynamics remain non-oscillatory (Section~\ref{sec:damping}).

\subsection{Muscle Responses as Second-Order Systems}
\label{sec:muscle_formulation}

Second-order linear systems have a long history as models of the muscular response to neural commands in speech. In the command--response model of Fujisaki \cite{fujisaki1984}, the fundamental-frequency contour is produced by laryngeal mechanisms driven by discrete phrase and accent commands, and each mechanism is modeled as a critically damped second-order linear system. This idea was later generalized to a superposition of muscle responses to spike commands, with trainable second-order filters embedded in a neural network \cite{marelli2019}.

The key observation underlying our readout is that the subthreshold dynamics of the AdLIF neuron already form such a system. In the absence of spikes, eliminating the adaptation variable $w$ from \eqref{eq:adlif_u}--\eqref{eq:adlif_w} yields a single second-order equation for the membrane potential,
\begin{equation}
	\tau_u \tau_w\, \ddot{u}(t) + (\tau_u + \tau_w)\, \dot{u}(t) + (1+a)\, u(t) = I(t) + \tau_w \dot{I}(t),
	\label{eq:secondorder}
\end{equation}
which is the equation of a damped mass--spring system, where the synaptic input current plays the role of an external driving force, i.e., the same class of spring--damper--mass models used for muscle responses \cite{marelli2019}. In a spiking neuron, this smooth second-order response remains hidden, because the threshold and reset convert it into discrete events. Obtaining a muscle model therefore requires no additional component and it suffices to remove the threshold, and to use the membrane potential itself as a continuous output. The spiking layer and its readout thus share a single neuron model with the same parameterization; one emits discrete events, while the other integrates them into a continuous trajectory.

In discrete time, removing the threshold corresponds to setting $s_t \equiv 0$ in \eqref{eq:adlif_disc_u}--\eqref{eq:adlif_disc_w}. The membrane potential $u_n(t)$ is updated as
\begin{align}
	u_n(t) = \alpha_n u_n(t-\Delta t) &+ (1-\alpha_n)\big(-w_n(t-\Delta t) \nonumber\\
	&+ x_n(t)\big),
	\label{eq:muscle_u}
\end{align}
where the first term represents a leaky memory of the previous state, controlled by $\alpha_n$, the second term introduces an inhibitory contribution through the adaptive variable $w_n$, and the third term corresponds to the external input $x_n(t)$. As a result, the membrane potential integrates past activity, adaptive suppression, and current input into a unified continuous state. The adaptive variable $w_n(t)$ evolves according to
\begin{equation}
	w_n(t) = \beta_n w_n(t-\Delta t) + (1-\beta_n)\,a_n u_n(t-\Delta t),
	\label{eq:muscle_w}
\end{equation}
where $\beta_n$ determines the time scale of the adaptive dynamics and $a_n$ controls its coupling to the membrane state. As in \eqref{eq:alphabeta}, $\alpha_n = \exp(-\Delta t/\tau_{u,n})$ and $\beta_n = \exp(-\Delta t/\tau_{w,n})$, where $\tau_{u,n}$ and $\tau_{w,n}$ are learned per neuron. The parameters are trainable and only constrained to predefined ranges, rather than being fixed.

Unlike standard spiking neurons, this module does not include a thresholding mechanism and therefore does not generate spikes. Since no thresholding or spike generation is applied, the continuous membrane potential $u_n(t)$ is used directly as the output.

\subsection{Second-Order Dynamics and Damping Regime}
\label{sec:damping}

The interaction between $u_n(t)$ and the slower adaptive state $w_n(t)$ produces the second-order temporal response of the system. To make this explicit, we collect the two states into $\mathbf{z}_n(t) = [u_n(t),\, w_n(t)]^\top$. Equations \eqref{eq:muscle_u}--\eqref{eq:muscle_w} then form a linear state-space system,
\begin{equation}
	\mathbf{z}_n(t) = \mathbf{A}_n\, \mathbf{z}_n(t-\Delta t) + \mathbf{B}_n\, x_n(t),
	\label{eq:statespace}
\end{equation}
with
\begin{equation}
	\mathbf{A}_n =
	\begin{bmatrix}
		\alpha_n & -(1-\alpha_n) \\
		(1-\beta_n)\,a_n & \beta_n
	\end{bmatrix},
	\quad
	\mathbf{B}_n =
	\begin{bmatrix}
		1-\alpha_n \\ 0
	\end{bmatrix}.
	\label{eq:AB}
\end{equation}

Because $\mathbf{A}_n$ is a $2\times 2$ matrix, each neuron acts as a second-order filter.
In contrast to a single leaky integrator (a first-order filter), whose impulse response consists of an exponential decay, the coupled second-order dynamics can produce smoother, rise-and-decay responses through the interaction of two temporal states.

The eigenvalues $\lambda_{1,2}$ of $\mathbf{A}_n$ are the roots of the characteristic equation
\begin{equation}
	\lambda^2 - \operatorname{tr}(\mathbf{A}_n)\,\lambda + \det(\mathbf{A}_n) = 0.
	\label{eq:charpoly}
\end{equation}
 Their nature is determined by the discriminant
\begin{align}
	\Delta_n &= \operatorname{tr}(\mathbf{A}_n)^2 - 4\det(\mathbf{A}_n) \nonumber\\
	&= (\alpha_n - \beta_n)^2 - 4\,a_n(1-\alpha_n)(1-\beta_n).
	\label{eq:discriminant}
\end{align}
Three regimes follow:
\begin{itemize}
	\item $\Delta_n > 0$ (\emph{overdamped}): $\lambda_{1,2}$ are real and distinct, and the impulse response is a weighted sum of two decaying exponentials, $\lambda_1^k$ and $\lambda_2^k$, giving a single rise-and-decay peak (a double-exponential kernel). Since $\operatorname{tr}(\mathbf{A}_n) > 0$, both eigenvalues are positive whenever $\det(\mathbf{A}_n) > 0$, so the response does not alternate in sign.

	\item $\Delta_n = 0$ (\emph{critically damped}): a repeated real eigenvalue, giving the fastest non-oscillatory response.
	\item $\Delta_n < 0$ (\emph{underdamped}): $\lambda_{1,2}$ are complex conjugates, and each spike produces a decaying oscillation.
\end{itemize}

Equation \eqref{eq:discriminant} also gives an intuitive reading of the overdamped condition. Since $(1-\alpha_n)(1-\beta_n) > 0$, the system is overdamped if and only if
\begin{equation}
	a_n < \frac{(\alpha_n - \beta_n)^2}{4\,(1-\alpha_n)(1-\beta_n)}.
	\label{eq:overdamped_cond}
\end{equation}
Thus, non-oscillatory behavior arises when the coupling $a_n$ between the two states is weak relative to the separation of their timescales. In particular, any neuron with $a_n \le 0$ is always overdamped, while oscillations require strong positive coupling between states with similar timescales.

For spike-to-continuous conversion, non-oscillatory responses are desirable, as the readout should integrate discrete spike events into smooth continuous trajectories without introducing additional oscillations. Both overdamped and critically damped systems can provide such responses, whereas underdamped dynamics
introduce oscillatory components that may distort the reconstructed signal. We do not impose $\Delta_n > 0$ during training; consequently, $\Delta_n$ and the resulting damping regime emerge from the learned parameters. The dynamical parameters are learned within predefined ranges, allowing us to investigate whether non-oscillatory, muscle-like dynamics emerge naturally from the speech reconstruction objective.

\section{Experimental Setup and Results}
\subsection{Dataset and Features}  
 To investigate this representation problem, we selected FastSpeech2\cite{fastspeech2} implemented with the SpeechBrain toolkit
 \cite{speechbrain} as a strong speech-synthesis baseline, enabling direct comparison with a text-to-speech (TTS) system. Speech synthesis provides an ideal benchmark because it requires accurate reconstruction of highly structured continuous signals while preserving fine temporal dependencies, making it particularly suitable for investigating temporal processing in spiking neural networks. Experiments were conducted on the LJSpeech dataset, which contains approximately 24 hours of single-speaker English speech comprising 13,100 audio recordings and their corresponding transcripts with 1,310 utterances reserved for evaluation. The audio recordings are sampled at 22.05 kHz and were represented using 80-dimensional mel-spectrograms for training and evaluation. Mel-spectrograms were extracted using an FFT frame size of 1024 and a frame period of 256 samples. The SPARSE framework \cite{bittar2022surrogate} was used to implement the spiking
layers. 
\subsection{Training}
\label{sec:training}

In all experiments, the pretrained FastSpeech2 model, including its decoder and vocoder, is kept frozen. Only the parameters of the spike-to-continuous pathway are optimized: the weights and neuron parameters of the AdLIF layer, and the readout weights $W_{nj}$ together with the dynamical parameters $(\alpha_n, \beta_n, a_n)$ of the muscle-response model. The pathway is trained to reconstruct the mel-spectrogram produced by the frozen FastSpeech2 decoder  and is optimized
for 100 epochs using Adam optimizer with a learning rate of $5 \times 10^{-4}$, a weight decay of $10^{-6}$, and a batch size of 48.  Because FastSpeech2 is frozen, any difference in output quality can be attributed solely to the spiking bottleneck.

\subsection{System Configuration}          
\begin{figure*}[!t]
	\centering
	\includegraphics[width=\textwidth]{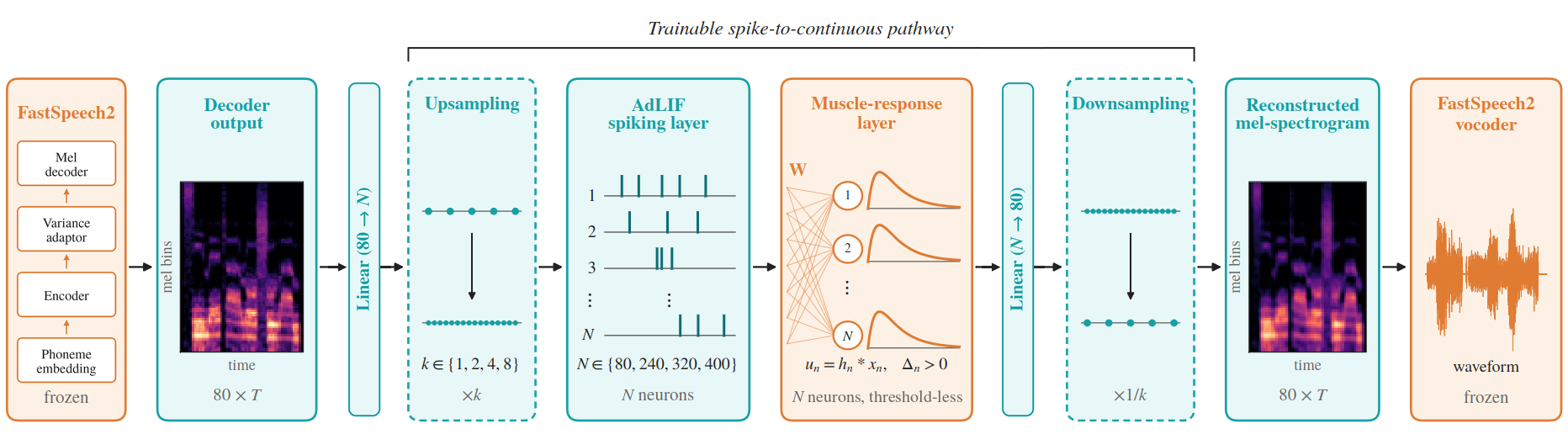}
	\caption{Overview of the proposed architecture. Linear layers map the 80 mel channels
		to the $N$ spiking neurons and the $N$ muscle-response outputs back to 80
		channels. Dashed boxes indicate the additional upsampling
		and downsampling operations used to investigate
		the effect of temporal resolution.}
	\label{fig:architecture}
\end{figure*}
As illustrated in Fig.~\ref{fig:architecture}, an AdLIF spiking
layer followed by the proposed muscle-response model was inserted between
the FastSpeech2 decoder and vocoder. This configuration enables direct
comparison between the original decoder representation and the
representation reconstructed after the spiking transformation. In the initial experiment to check the general possibility of using membrane potential as a smooth output, both the spiking and muscle-response
layers contained 80 neurons, matching the 80-channel mel-spectrogram
produced by the FastSpeech2 decoder. 

\begin{figure}[bhtp]
	\centering
	\includegraphics[width=\columnwidth]{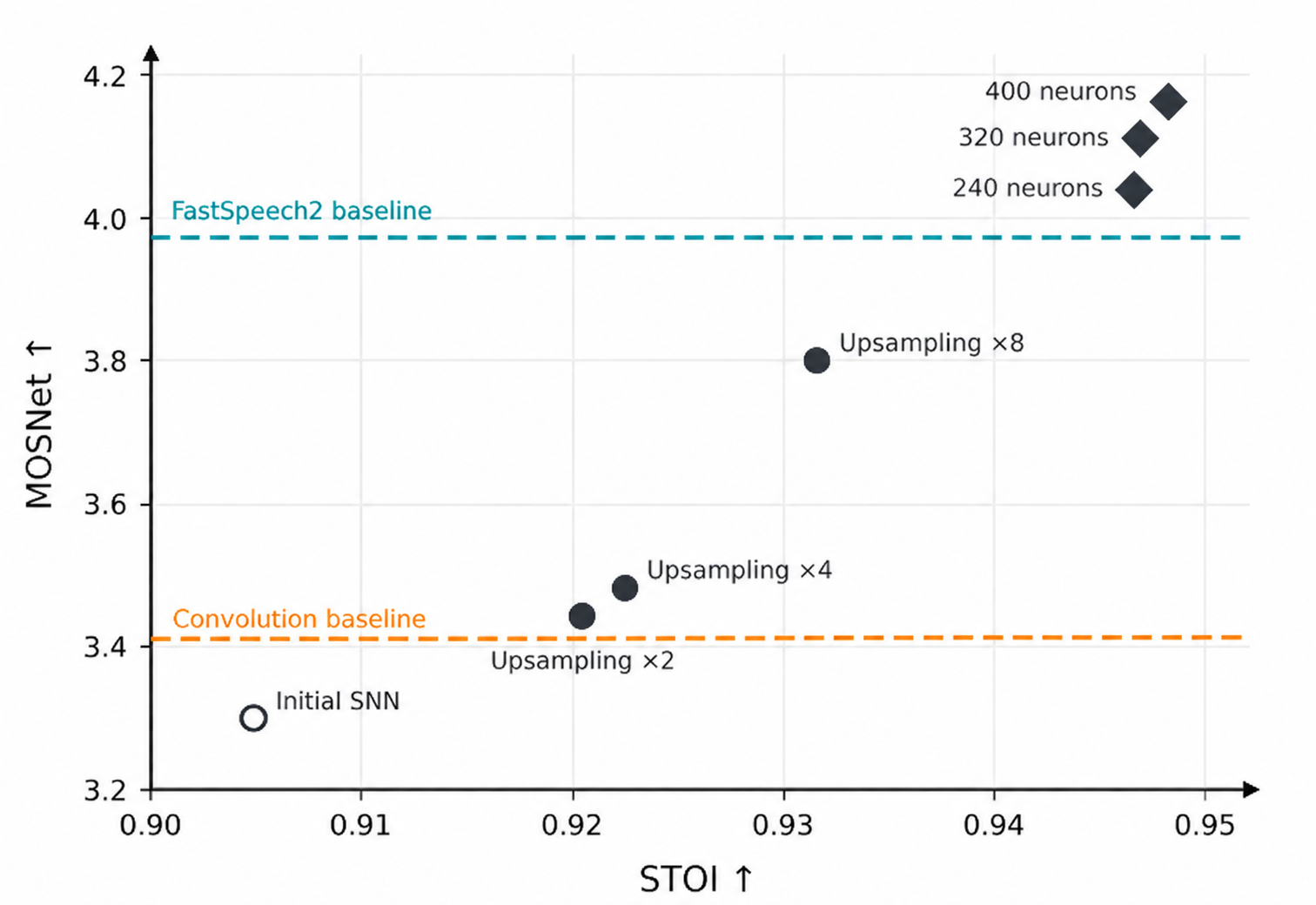}
	\caption{MOSNet and STOI performance across different spike-to-continuous configurations. Dashed lines indicate the FastSpeech2 and convolutional-readout baselines.}
	\label{fig:mosnet}
\end{figure}

 As shown in Fig.~\ref{fig:mosnet}, the initial spiking
configuration already produced intelligible speech (STOI $\approx 0.90$),
supporting our hypothesis that SNNs can be used for
continuous acoustic reconstruction. However, the
substantial degradation in reconstruction quality
compared with the non-spiking baseline indicates that
the initial configuration is not sufficient to preserve
acoustic information with high fidelity.

This observation revealed two key design constraints, temporal resolution and representational capacity, which form the focus of the following investigation. To isolate the effect of each factor, temporal resolution
and representational capacity were investigated independently,
with the other factor held constant in each experiment.
This experimental design allows us to assess their
individual contributions to reconstruction quality.  Our first design question is, does increasing temporal resolution facilitate spike-to-continuous conversion? The second question is whether increasing the number
of spiking neurons, and thus the representational
capacity of the spiking layer, improves the quality
of the reconstructed acoustic features.

The first question was inspired by the oversampling principle of sigma-delta digital-to-analog converters. 
We expected that increasing temporal resolution before
spike generation would provide more opportunities for
spike-based encoding of the acoustic features, thereby
improving mel-spectrogram reconstruction quality.
 So, in the first part of the experiments as shown in Fig. \ref{fig:architecture}, we increased the temporal resolution before spike generation by linearly interpolating the mel-spectrogram along the time axis,
 using upsampling factors of 2, 4, and 8. For the second question, we further expected that the initial 80-neuron spiking layer was insufficient to faithfully encode the rich temporal and spectral structure of speech, motivating the investigation of larger spiking layers with greater representational capacity and in part of the experiments, we increased the number of spiking neurons.

To assess whether the proposed muscle-response model provides an
advantage over a conventional learned readout, we additionally replaced
the muscle-model layer with a convolutional readout while keeping
the preceding spiking representation unchanged. The convolutional
baseline consisted of two one-dimensional convolutional layers with a
kernel size of 9 and same padding, separated by a Tanh activation.
This provides a direct comparison between the proposed dynamical
spike-to-continuous mechanism and a conventional trainable temporal
readout.

\subsection{Results of Evaluation Metrics}            
Fig. \ref{fig:mosnet} summarizes the performance of the proposed muscle-response model across different architectural configurations using two objective metrics: MOSNet \cite{mosnet}, which estimates perceived speech quality, and STOI \cite{stoi}, which assesses speech intelligibility relative to a reference signal. Here, the reference is the waveform synthesized by the vocoder from the non-spiking FastSpeech2 mel-spectrogram, so STOI measures how faithfully the spiking pathway preserves the FastSpeech2 output. MOSNet, in contrast, is reference-free. Higher values indicate better performance for both metrics. All configurations were evaluated on the same set of test utterances to ensure a consistent comparison across models. Since this study primarily investigates the feasibility of spike-based speech generation and the influence of architectural parameters on output quality, we rely on objective metrics for consistent comparisons across configurations, rather than conducting subjective listening tests. Introducing the spiking layer initially reduced the MOSNet score from 3.973 for FastSpeech2 to 3.300, demonstrating the difficulty of reconstructing continuous acoustic representations from spike trains. As an additional baseline, the proposed muscle-response layer was replaced with the convolutional readout described above. This conventional readout achieved a MOSNet score of 3.417. At the matched baseline configuration, the convolutional readout performs slightly better than the muscle-response model (3.417 vs. 3.300). However, with increased temporal resolution or capacity, the muscle-response pathway clearly exceeds it, while using a compact, interpretable, and biologically grounded readout. Evaluating the convolutional readout under the same enlarged configurations is left for future work.

By investigating the first design question, increasing the temporal resolution through 8× upsampling recovered most of the degradation (MOSNet = 3.801), highlighting the importance of temporal resolution for spike-based speech synthesis. A consistent improvement is also observed as the upsampling factor increases from $2\times$ to $4\times$ and $8\times$, confirming that finer temporal resolution facilitates spike-based representation of the acoustic features. 

Investigating the second design question showed that expanding the spiking layer from 80 to 400 neurons increased the MOSNet score to 4.163, surpassing the original FastSpeech2 baseline. Since the pathway is trained to reconstruct the FastSpeech2 mel-spectrogram, this should not be interpreted as an improvement over its target; rather, it indicates that the spiking bottleneck no longer degrades the estimated quality, while the small residual deviation from the reference (STOI $\approx$ 0.95) may reflect a mild smoothing effect that MOSNet favors. More importantly, performance improves consistently as the number of spiking neurons is increased from 80 to 240, 320, and 400. This trend indicates that representational capacity plays a stronger role than temporal oversampling in preserving the acoustic information encoded by the continuous representation. Interestingly, the variation in MOSNet scores is consistent with that observed in STOI. Consequently, the best-performing configurations are concentrated in the upper-right region, corresponding to simultaneously high MOSNet and STOI scores.

Motivated by the relatively small number of muscle groups involved in speech articulation, we also conducted preliminary experiments to investigate whether the muscle-response model could operate with a substantially reduced number of  neurons for muscle-response layer. However, reducing the number of neurons led to a degradation in speech quality, suggesting that the required model dimensionality cannot be directly inferred from the number of articulatory muscles in this step. A systematic investigation of this relationship is left for future work.

\subsection{Analysis of the Learned Dynamics}       
\begin{figure}[bhtp]
	\centering
	\includegraphics[width=\columnwidth]{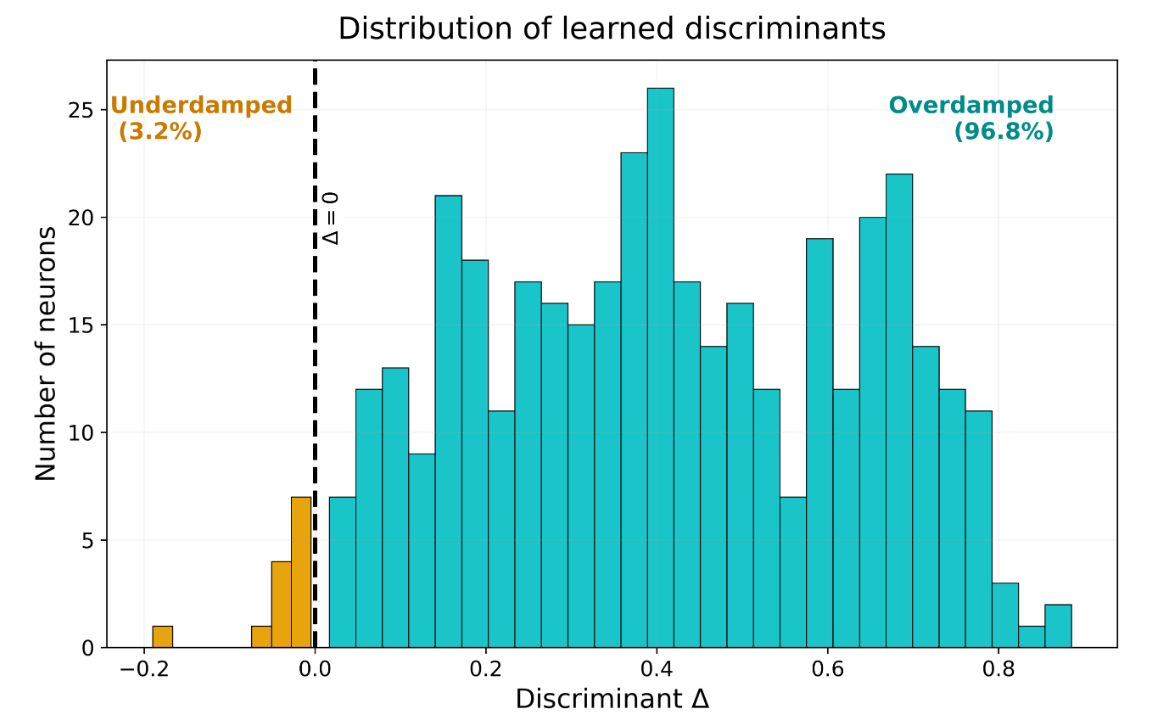}
	\caption{Distribution of muscle response neurons across different regimes for the 400-neuron configuration.}
	\label{fig:delta}
\end{figure}

Using the discriminant $\Delta_n$ in \eqref{eq:discriminant}, we examined the damping regime of each learned muscle-response neuron. Although no constraint on the damping regime was imposed during training, 96.8\% of the neurons fall within the overdamped region ($\Delta_n > 0$), some of them close to the critical-damping boundary (Fig.~\ref{fig:delta}). Each spike therefore evokes a smooth, single-peaked, double-exponential-like response, which is well suited to integrating sparse spike events into continuous representations and is consistent with the muscle-like behavior that motivated the model.

\begin{figure}[bhtp]
	\centering
	\includegraphics[width=\columnwidth]{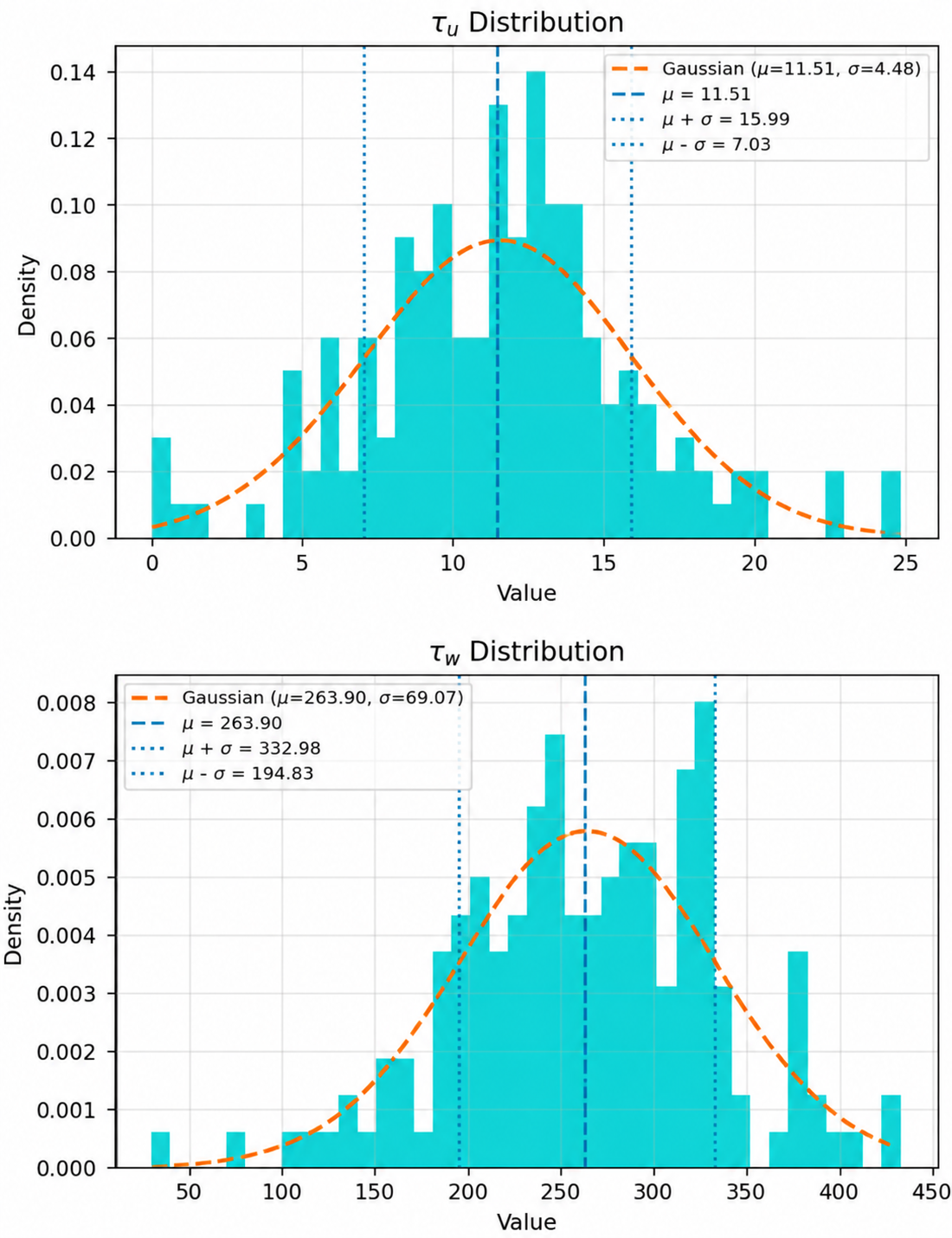}
	\caption{Distribution of the learned membrane and adaptation time constants of the muscle-response neurons}
	\label{fig:tau}
\end{figure}

After training, we also examined the distribution of the learned time constants (Fig.~\ref{fig:tau}). The learned membrane time constants are concentrated around $\tau_u \approx 11$ ms, close to the frame-level temporal resolution commonly used in speech processing, whereas the adaptation time constants are centered around $\tau_w \approx 264$ ms, approaching the timescale associated with syllabic structure. This suggests that the learned dynamics operate on speech-relevant timescales rather than arbitrary parameter values.

\section{Conclusion}

In this work, we investigated the problem of reconstructing continuous
representations from sparse spiking activity and proposed a biologically
inspired muscle-response model as a spike-to-continuous interface.
Using speech synthesis as a challenging continuous reconstruction task,
we showed that both temporal resolution and representational capacity
play important roles in preserving information through a spiking
bottleneck. Increasing temporal resolution progressively improved
reconstruction quality, while increasing the number of spiking neurons
produced a substantially larger improvement. Furthermore, the muscle-response readout, with only three dynamical parameters per neuron, outperformed a conventional convolutional readout evaluated at the baseline configuration once sufficient capacity or temporal resolution was provided.

Analysis of the learned muscle-response dynamics further showed that,
despite imposing no explicit constraint on the damping regime, the
large majority of neurons converged to the overdamped region, producing non-oscillatory responses.
These findings suggest that successful spike-to-continuous conversion
depends not only on the temporal resolution of spike trains, but also
critically on sufficient representational capacity and appropriate
continuous temporal dynamics. Overall, the proposed approach provides
a biologically motivated interface between discrete spiking computation
and continuous-valued representations, supporting the development of
spiking models for continuous and generative tasks.

\section*{Acknowledgment}
This work was supported by the Swiss National Science Foundation (SNSF)
under SpiDeR \href{https://data.snf.ch/grants/grant/10004827}{grant number 10004827}. The authors would like to thank Prof. Wulfram
Gerstner for his insightful comments and suggestions during an earlier
presentation of this research direction.

\IEEEtriggeratref{11}

\vspace{12pt}


\begin{thebibliography}{00}

	\bibitem{Gerstner_Kistler_2002}
	W.~Gerstner and W.~M.~Kistler,
	\textit{Spiking Neuron Models: Single Neurons, Populations, Plasticity}.
	Cambridge, U.K.: Cambridge University Press, 2002,
	doi: \href{https://doi.org/10.1017/CBO9780511815706}{10.1017/CBO9780511815706}.
	
	\bibitem{guenther2016speech}
	F.~H.~Guenther,
	\textit{Neural Control of Speech}.
	Cambridge, MA, USA: MIT Press, 2016.
	
	\bibitem{bouchard2013speech}
	K.~E.~Bouchard, N.~Mesgarani, K.~Johnson, and E.~F.~Chang,
	``Functional organization of human sensorimotor cortex for speech articulation,''
	\textit{Nature}, vol. 495, no. 7441, pp. 327--332, 2013,
	doi: \href{https://doi.org/10.1038/nature11911}{10.1038/nature11911}.
	

	\bibitem{heckman2012motorunit}
	C.~J.~Heckman and R.~M.~Enoka,
	``Motor unit,''
	\textit{Compr. Physiol.}, vol. 2, no. 4, pp. 2629--2682, 2012,
	doi: \href{https://doi.org/10.1002/cphy.c100087}{10.1002/cphy.c100087}.
	
	\bibitem{fang2021plif}
	W.~Fang, Z.~Yu, Y.~Chen, T.~Masquelier, T.~Huang, and Y.~Tian,
	``Incorporating learnable membrane time constant to enhance learning of spiking neural networks,''
	in \textit{Proc. IEEE/CVF Int. Conf. Computer Vision (ICCV)}, 2021, pp. 2661--2671,
	doi: \href{https://doi.org/10.1109/ICCV48922.2021.00266}{10.1109/ICCV48922.2021.00266}.
	
	\bibitem{bittar2022surrogate}
	A.~Bittar and P.~N.~Garner,
	``A surrogate gradient spiking baseline for speech command recognition,''
	\textit{Front. Neurosci.}, vol. 16, Art. no. 865897, 2022,
	doi: \href{https://doi.org/10.3389/fnins.2022.865897}{10.3389/fnins.2022.865897}.
	
	\bibitem{bittar2022}
	A.~Bittar and P.~N.~Garner,
	``Surrogate gradient spiking neural networks as encoders for large vocabulary continuous speech recognition,''
	arXiv preprint arXiv:2212.01187, 2022,
	doi: \href{https://doi.org/10.48550/arXiv.2212.01187}{10.48550/arXiv.2212.01187}.
	
	\bibitem{neuromorphiccomputing}
	K.~Roy, A.~Jaiswal, and P.~Panda,
	``Towards spike-based machine intelligence with neuromorphic computing,''
	\textit{Nature}, vol. 575, no. 7784, pp. 607--617, 2019,
	doi: \href{https://doi.org/10.1038/s41586-019-1677-2}{10.1038/s41586-019-1677-2}.
	
	\bibitem{fono2026mathematical}
	A.~Fono, M.~Singh, E.~Araya, P.~C.~Petersen, H.~Boche, and G.~Kutyniok,
	``Mathematical foundations of spiking neural networks: Strengths, challenges, and computational paradigm potential,''
	\textit{IEEE Signal Process. Mag.}, vol. 43, no. 2, pp. 64--76, Mar. 2026,
	doi: \href{https://doi.org/10.1109/MSP.2025.3597033}{10.1109/MSP.2025.3597033}.
	
	\bibitem{spikevoice}
	K.~Wang, J.~Zhang, Y.~Ren, M.~Yao, D.~Shang, B.~Xu, and G.~Li,
	``SpikeVoice: High-quality text-to-speech via efficient spiking neural network,''
	in \textit{Proc. 62nd Annu. Meeting Assoc. Comput. Linguistics (ACL), Vol. 1: Long Papers},
	2024, pp. 7927--7940,
	doi: \href{https://doi.org/10.18653/v1/2024.acl-long.429}{10.18653/v1/2024.acl-long.429}.
	
	\bibitem{SRM}
	R.~Jolivet, T.~J.~Lewis, and W.~Gerstner,
	``The spike response model: A framework to predict neuronal spike trains,''
	in \textit{Artificial Neural Networks and Neural Information Processing -- ICANN/ICONIP 2003}, ser. Lecture Notes in Computer Science, vol. 2714.
	Berlin, Germany: Springer, 2003, pp. 846--853,
	doi: \href{https://doi.org/10.1007/3-540-44989-2_101}{10.1007/3-540-44989-2\_101}.
	
	\bibitem{marelli2019}
	F.~Marelli, B.~Schnell, H.~Bourlard, T.~Dutoit, and P.~N.~Garner,
	``An end-to-end network to synthesize intonation using a generalized command response model,''
	in \textit{Proc. IEEE Int. Conf. Acoust., Speech Signal Process. (ICASSP)},
	2019, pp. 7040--7044,
	doi: \href{https://doi.org/10.1109/ICASSP.2019.8683815}{10.1109/ICASSP.2019.8683815}.
	
	\bibitem{fastspeech2}
	Y.~Ren, C.~Hu, X.~Tan, T.~Qin, S.~Zhao, Z.~Zhao, and T.-Y.~Liu,
	``FastSpeech 2: Fast and high-quality end-to-end text to speech,''
	in \textit{Proc. Int. Conf. Learning Representations (ICLR)}, 2021.
	
	\bibitem{majumdar2025spiking}
	P.~Majumdar,
	``Spiking neural networks: A comprehensive review of diverse applications,
	research progress, challenges and future research directions,''
	\textit{Evolving Syst.}, vol. 16, no. 4, Art. no. 125, 2025,
	doi: \href{https://doi.org/10.1007/s12530-025-09755-0}{10.1007/s12530-025-09755-0}.
	
	\bibitem{lapicque1907recherches}
	L.~Lapicque,
	``Recherches quantitatives sur l'excitation \'electrique des nerfs trait\'ee comme une polarisation,''
	\textit{J. Physiol. Pathol. G\'en.}, vol. 9, pp. 620--635, 1907.
	
	\bibitem{baronig2025}
	M.~Baronig, R.~Ferrand, S.~Sabathiel, and R.~Legenstein,
	``Advancing spatio-temporal processing through adaptation in spiking neural networks,''
	\textit{Nature Commun.}, vol. 16, Art. no. 5776, 2025,
	doi: \href{https://doi.org/10.1038/s41467-025-60878-z}{10.1038/s41467-025-60878-z}.
	
	\bibitem{hodgkin}
	A.~L.~Hodgkin and A.~F.~Huxley,
	``A quantitative description of membrane current and its application to conduction and excitation in nerve,''
	\textit{J. Physiol.}, vol. 117, no. 4, pp. 500--544, 1952,
	doi: \href{https://doi.org/10.1113/jphysiol.1952.sp004764}{10.1113/jphysiol.1952.sp004764}.
	
	\bibitem{sparse}
	A.~Bittar and P.~N.~Garner,
	``Exploring neural oscillations during speech perception via surrogate gradient spiking neural networks,''
	\textit{Front. Neurosci.}, vol. 18, Art. no. 1449181, 2024,
	doi: \href{https://doi.org/10.3389/fnins.2024.1449181}{10.3389/fnins.2024.1449181}.
	
	\bibitem{augustin}
	M.~Augustin, J.~Ladenbauer, and K.~Obermayer,
	``How adaptation shapes spike rate oscillations in recurrent neuronal networks,''
	\textit{Front. Comput. Neurosci.}, vol. 7, Art. no. 9, 2013,
	doi: \href{https://doi.org/10.3389/fncom.2013.00009}{10.3389/fncom.2013.00009}.
	
	\bibitem{neftci2019surrogate}
	E.~O.~Neftci, H.~Mostafa, and F.~Zenke,
	``Surrogate gradient learning in spiking neural networks,''
	\textit{IEEE Signal Process. Mag.}, vol. 36, no. 6, pp. 51--63, 2019,
	doi: \href{https://doi.org/10.1109/MSP.2019.2931595}{10.1109/MSP.2019.2931595}.
	
	
	\bibitem{fujisaki1984}
	H.~Fujisaki and K.~Hirose,
	``Analysis of voice fundamental frequency contours for declarative sentences of Japanese,''
	\textit{J. Acoust. Soc. Jpn. (E)}, vol. 5, no. 4, pp. 233--242, 1984,
	doi: \href{https://doi.org/10.1250/ast.5.233}{10.1250/ast.5.233}.
	
	\bibitem{speechbrain}
	M.~Ravanelli, T.~Parcollet, A.~Moumen, S.~de~Langen, C.~Subakan,
	P.~Plantinga, Y.~Wang, P.~Mousavi, L.~Della~Libera, A.~Ploujnikov,
	F.~Paissan, D.~Borra, S.~Zaiem, Z.~Zhao, S.~Zhang, G.~Karakasidis,
	S.-L.~Yeh, P.~Champion, A.~Rouhe, R.~Braun, F.~Mai, J.~Zuluaga-Gomez,
	S.~M.~Mousavi, A.~Nautsch, H.~Nguyen, X.~Liu, S.~Sagar, J.~Duret,
	S.~Mdhaffar, G.~Laperri\`ere, M.~Rouvier, R.~De~Mori, and Y.~Est\`eve,
	``Open-source conversational AI with SpeechBrain 1.0,''
	\textit{J. Mach. Learn. Res.}, vol. 25, no. 333, pp. 1--11, 2024.
	
	\bibitem{mosnet}
	C.-C.~Lo, S.-W.~Fu, W.-C.~Huang, X.~Wang, J.~Yamagishi, Y.~Tsao, and H.-M.~Wang,
	``MOSNet: Deep learning-based objective assessment for voice conversion,''
	in \textit{Proc. Interspeech}, 2019, pp. 1541--1545,
	doi: \href{https://doi.org/10.21437/Interspeech.2019-2003}{10.21437/Interspeech.2019-2003}.
	
	\bibitem{stoi}
	C.~H.~Taal, R.~C.~Hendriks, R.~Heusdens, and J.~Jensen,
	``An algorithm for intelligibility prediction of time--frequency weighted noisy speech,''
	\textit{IEEE Trans. Audio, Speech, Lang. Process.}, vol. 19, no. 7, pp. 2125--2136, 2011,
	doi: \href{https://doi.org/10.1109/TASL.2011.2114881}{10.1109/TASL.2011.2114881}.
	
\end{thebibliography}
\end{document}